\documentclass{article} 
\usepackage{iclr2026_conference,times}

\usepackage{amsmath,amsfonts,bm}

\def\eqref#1{equation~\ref{#1}}

\def\1{\bm{1}}

\DeclareMathAlphabet{\mathsfit}{\encodingdefault}{\sfdefault}{m}{sl}
\SetMathAlphabet{\mathsfit}{bold}{\encodingdefault}{\sfdefault}{bx}{n}

\usepackage{hyperref}
\usepackage{url}
\usepackage{graphicx}
\usepackage{xspace}
\usepackage{colortbl}
\usepackage{booktabs}
\usepackage{multirow,marvosym}
\usepackage{wrapfig}

\usepackage{xcolor}

\definecolor{ForestGreen}{rgb}{0, 0.69, 0.31}
\definecolor{NavyBlue}{rgb}{0, 0.44, 0.75}
\newcommand{\hgreen}[1]{\textcolor{ForestGreen}{\textbf{#1}}} 

\usepackage{pifont}

\definecolor{00blue}{RGB}{139,169,235}

\usepackage[capitalise]{cleveref}
\newcommand{\methodname}{SPARK\xspace}
\newcommand{\sparkvlsmall}{SPARK-VL-7B\xspace}
\newcommand{\sparkvlbig}{SPARK-VL-32B\xspace}
\newcommand{\sparkllm}{SPARK-7B\xspace}

\definecolor{Cerulean}{rgb}{0.0, 0.48, 0.65}

\definecolor{customblue}{HTML}{005AD7}

\usepackage{mathtools}
\DeclarePairedDelimiter\set\{\}

\title{
\vspace{-2em}
  \hspace{-1em}
  \raisebox{-2.5ex}{%
    \protect\includegraphics[height=4.0\fontcharht\font`\B]{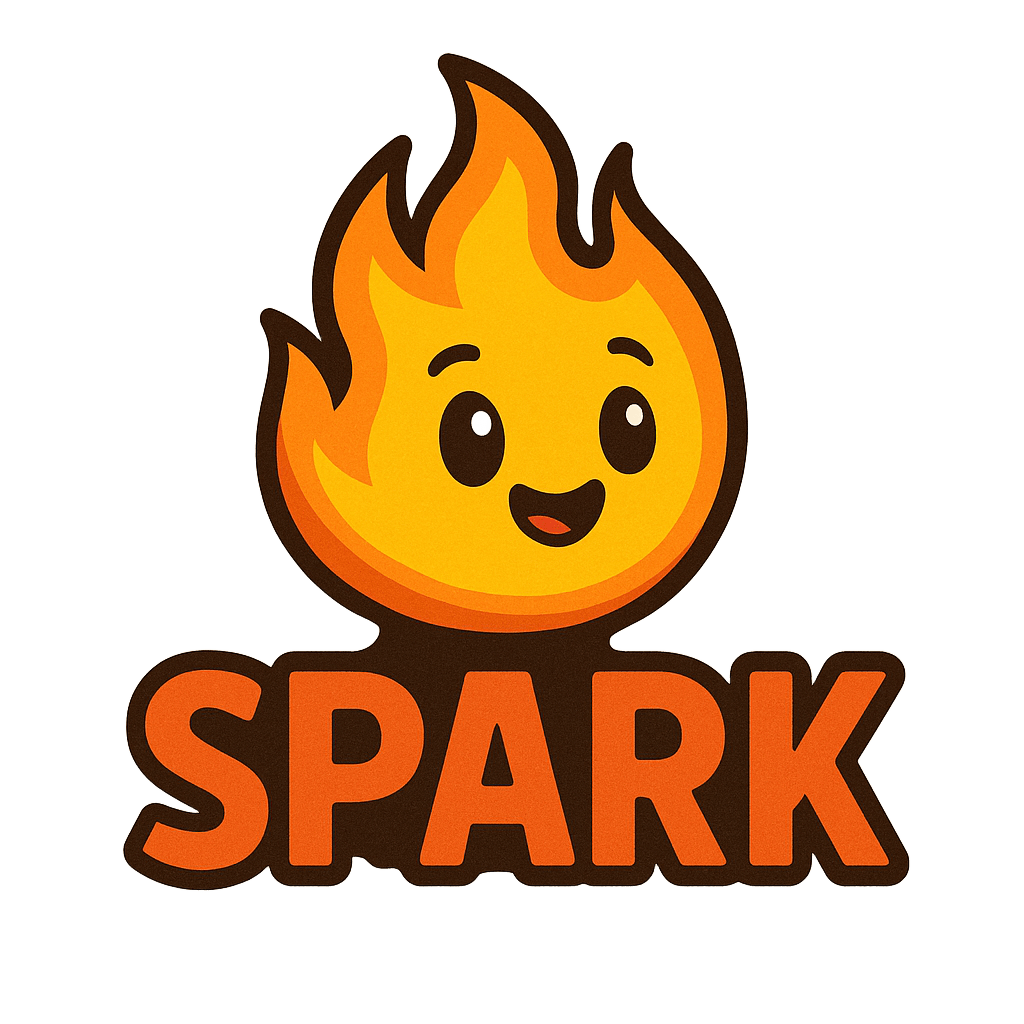}%
  }%
SPARK: Synergistic Policy And Reward co-evolving framework}

\author{Ziyu Liu$^{1,2}$, Yuhang Zang$^{2}$\textsuperscript{\Letter}, Shengyuan Ding$^{2,3}$, Yuhang Cao$^{2}$, Xiaoyi Dong$^{2,4}$, \\
\textbf{Haodong Duan$^{2}$, Dahua Lin$^{2,4}$, Jiaqi Wang$^{2,5}$\textsuperscript{\Letter}} \\
$^{1}$Shanghai Jiao Tong University \quad $^{2}$Shanghai Artificial Intelligence Laboratory \\
$^3$Fudan University \quad $^4$The Chinese University of Hong Kong \quad $^5$Shanghai Innovation Institute \\
\texttt{liuziyu77@sjtu.edu.cn, zangyuhang@pjlab.org.cn} \\
{{\tt\small \textbf{Model \& Data}: \href{https://huggingface.co/collections/laolao77/spark-collection-68d67a15ef216bfd90892c99}{\textcolor{customblue}{Spark HuggingFace Collection}}}} \\
  {{\tt\small \textbf{Code}: \href{https://github.com/InternLM/Spark}{\textcolor{customblue}{Spark Github Repository}}}} 
}

\iclrfinalcopy 
\begin{document}

\maketitle

\begin{abstract}
Recent Large Language Models (LLMs) and Large Vision-Language Models (LVLMs) increasingly use Reinforcement Learning (RL) for post-pretraining, such as RL with Verifiable Rewards (RLVR) for objective tasks and RL from Human Feedback (RLHF) for subjective tasks.
However, RLHF incurs high costs and potential reward–policy mismatch due to reliance on human preferences, while RLVR still wastes supervision by discarding rollouts and correctness signals after each update.
To address these challenges, we introduce the \textbf{S}ynergistic \textbf{P}olicy \textbf{A}nd \textbf{R}eward Co-Evolving Framewor\textbf{K} (SPARK), an efficient, on-policy, and stable method that builds on RLVR.
Instead of discarding rollouts and correctness data, SPARK recycles this valuable information to simultaneously train the model itself as a generative reward model.
This auxiliary training uses a mix of objectives, such as pointwise reward score, pairwise comparison, and evaluation conditioned on further-reflection responses, to teach the model to evaluate and improve its own responses.
Our process eliminates the need for a separate reward model and costly human preference data.
SPARK creates a positive co-evolving feedback loop: improved reward accuracy yields better policy gradients, which in turn produce higher-quality rollouts that further refine the reward model.
Our unified framework supports test-time scaling via self-reflection without external reward models and their associated costs.
We show that SPARK achieves significant performance gains on multiple LLM and LVLM models and multiple reasoning, reward models, and general benchmarks.
For example, SPARK-VL-7B achieves an average 9.7\% gain on 7 reasoning benchmarks, 12.1\% on 2 reward benchmarks, and 1.5\% on 8 general benchmarks over the baselines, demonstrating robustness and broad generalization.
\end{abstract}
\section{Introduction}

Reinforcement learning (RL) is a standard step of post‑pretraining improvement and alignment for Large Language Models (LLMs) and Large Vision‑Language Models (LVLMs).
In practice, current RL systems rely on two complementary routes:
\textbf{(1)} RL with verifiable rewards (\textbf{RLVR}) \citep{lambert2024t,DeepSeek-R1,team2025kimi}, which uses a verifier to address objective and verifiable problems like math and code.
\textbf{(2)} Reward‑model–based pipelines such as RL from Human Feedback (\textbf{RLHF}) \citep{ouyang2022training,bai2022constitutional}, which distill human or synthetic preferences into a learned reward model to guide policy optimization on subjective tasks.
These two RL stages have yielded significant gains in reasoning quality, safety, and truthfulness, and have become a cornerstone in modern LLM/LVLM training.

Despite impressive progress, current RL pipelines for LLMs/LVLMs still exhibit several limitations.
Approaches based on verifiable rewards (RLVR) are effective only for tasks with explicit verifiers, leaving open-ended objectives like helpfulness and safety unaddressed.
Conversely, reward‑model–based pipelines (RLHF) can handle subjective tasks with reward models \citep{su2025crossing} or LLM-as-a-judge \citep{zheng2023judging,gunjal2025rubrics} but demand substantial and costly curated human preference data.
Furthermore, training the reward model as a separate component causes it to lag the evolving policy, inducing reward-policy mismatch, reward hacking, and brittle generalization under out-of-distribution queries \citep{skalse2022defining,gao2023scaling}.
Finally, dependence on external reward models or judge models introduces significant latency and serving costs during both training and test-time scaling \citep{zhao2025genprm}.

\begin{figure}[t]
    \begin{center}
    \includegraphics[width=.98\linewidth]{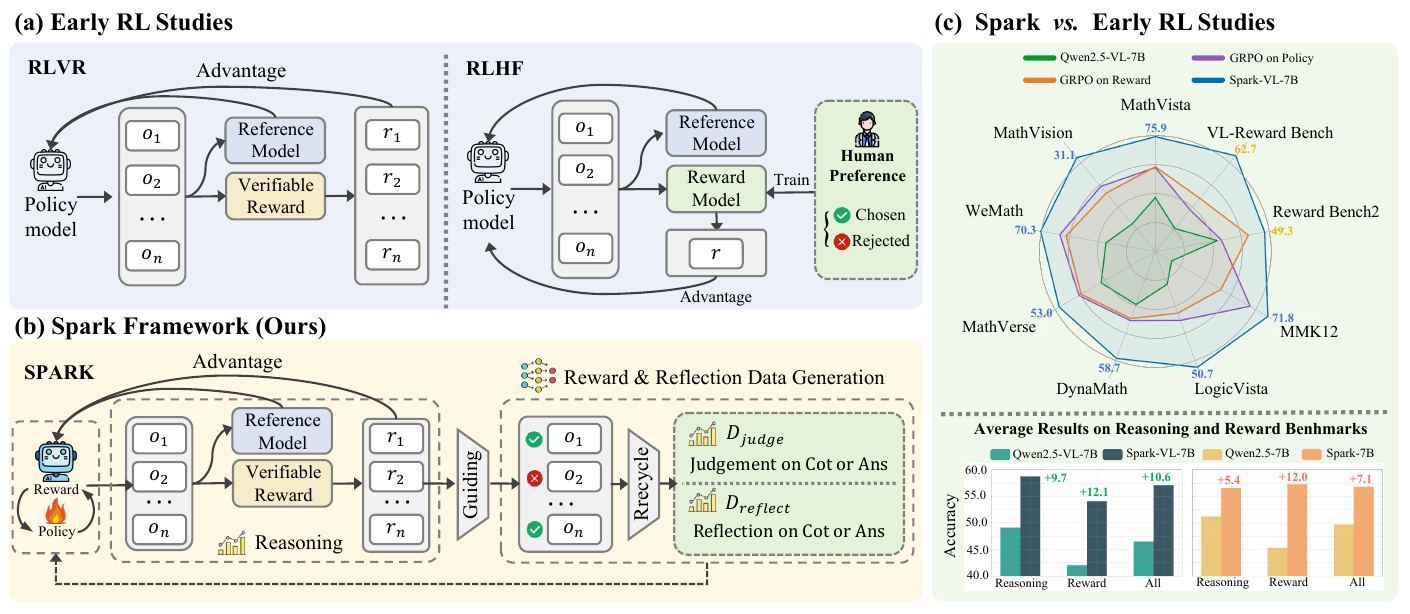}
    \end{center}
    \vspace{-16pt}
    \caption{\small 
    \textbf{(a)} Early studies of RL with Verifiable Rewards (RLVR) and RL from Human Feedback (RLHF) that rely on reward models.
    \textbf{(b)} We propose \methodname that recycles the rollouts from the RLVR, and further trains the model itself as a generative reward model. 
    \textbf{(c)} \methodname consistently outperforms early RL approaches in both reasoning and reward model benchmarks.
    }
    \vspace{-24pt}
    \label{fig:teaser}
\end{figure}

To mitigate the limitations of early RL studies, such as costs of human preference labeling and deployment, we turn to an \textit{internalized} source of supervision.
Our method builds on RL with Verifiable Rewards (RLVR), where $n$ candidate responses or rollouts $\{o_{1}, o_{2}, \ldots, o_{n}\}$ are generated, and score them against a ground-truth label to update the policy model (see \cref{fig:teaser} \textbf{(a)}).
However, these valuable rollouts are typically \textbf{discarded} after this single use.
Our key insight is to \textbf{recycle} the rollouts and correctness data to further train the model itself as a generative reward model simultaneously.
We use the RLVR-derived correctness scores to train the model on a mix of objectives: a pointwise objective to determine if a response is correct, a pairwise objective to identify which response is better, and a reflection objective to learn how to fix an incorrect response to get the correct one.

The proposed \textit{auxiliary} training paradigm for RLVR, the \textbf{S}ynergistic \textbf{P}olicy \textbf{A}nd \textbf{R}eward Co-Evolving Framewor\textbf{K} (\textbf{\methodname}), enhances reward accuracy, yielding stronger policy gradients and improving the model's reasoning abilities (see \cref{fig:teaser} \textbf{(b)}).
We further use this internal judge for self-reflection at test time, extending alignment to tasks beyond strictly verifiable domains while retaining the robustness of verifiable feedback.
\methodname has four advantages: (1) \textbf{Data- and compute-efficient}: no extra human preference data annotation or separate reward model training loop is required, as the signals come ``for free'' from RLVR training rollouts.
(2) \textbf{On-policy and stable}: reward data are continually sampled from and calibrated to the model’s current behavior, reducing reward–policy mismatch.
(3) \textbf{Co-evolving}: improved reward accuracy yields better gradients for the policy, which produces high-quality rollouts, further refining the reward.
(4) \textbf{Unified development}: our framework enables RL training and test‑time scaling, removing the dependency on an external reward model, and thereby saving GPU memory and reducing the communication overhead.

Our \methodname is applicable to both LLMs (e.g., Qwen2.5 \citep{yang2025qwen3}) and LVLMs (e.g., Qwen2.5-VL \citep{bai2025qwen25vl}).
As shown on \cref{fig:teaser} \textbf{(c)}, \methodname achieves clear improvements on various mathematical reasoning and reward model benchmarks.
For LVLMs, \sparkvlsmall improves by \hgreen{9.7\%} on 7 reasoning and \hgreen{12.1\%} on 2 reward benchmarks, in addition to an average \hgreen{1.5\%} gain on 8 general benchmarks.
These improvements are observed also with larger LVLM models (\sparkvlbig) and pure LLMs (\sparkllm), demonstrating the robustness across different model scales and architectures.

Our key contributions are:
\textbf{(1)} We introduce an efficient, on-policy, and stable framework \methodname that builds on RLVR but recycles the valuable rollouts that are typically discarded after policy updates.
We use the RLVR-derived correctness scores to train the model itself to become a generative reward model, which eliminates the need for human preference data to train a separate, external reward model with additional development costs.
\textbf{(2)} Our SPARK is designed as a \textbf{co-evolving} mechanism.
Improved reward accuracy yields better gradients for the policy, which in turn produces higher-quality rollouts. These high-quality rollouts further refine the reward model, creating a positive feedback loop that leads to stronger overall performance and stability.
\textbf{(3)} Extensive experiments show that \methodname achieves substantial improvements on multiple LVLM and LLM models.
\sparkvlsmall achieves average \hgreen{9.7\%} gains on 7 reasoning benchmarks, \hgreen{12.1\%} on 2 reward benchmarks, and \hgreen{1.5\%} on 8 general benchmarks, demonstrating the strong generalization ability.

\section{Related Works}

\noindent \textbf{Reinforcement Learning with Verifiable Reward.}
Following the success of DeepSeek-R1~\citep{DeepSeek-R1}, the GRPO~\citep{grpo} algorithm—driven by verifiable rewards—has demonstrated strong potential across a variety of reasoning-intensive tasks, particularly in mathematics and programming. Moreover, this RLVR paradigm has been successfully extended to a wide range of domains, including perception~\citep{zheng2025deepeyes,su2025openthinkimg, liu2025visualrft, peng2025lmmr1}, agent~\citep{jin2025searchr1, liu2025visualarft} and so on. In this work, we adopt a GRPO-based algorithm to build our synergistic policy and reward co-evolving framework. Through reinforcement learning, our framework jointly enhances the policy’s reasoning and the reward’s judging abilities in a unified model, breaking the isolation between policy and reward models in prior approaches.

\noindent \textbf{Reinforcement Learning from Human Feedback.} 
Reinforcement Learning from Human Feedback (RLHF) optimizes policy models using human preference data. 
These data are either directly collected from human annotations or generated by teacher models, and are typically used to first train an independent reward model. 
The reward model then provides feedback signals that guide policy optimization~\citep{Cai2024InternLM2TR, starling2023, xcomposer2.5-reward, Kim2023SOLAR1S, Yuan2024SelfRewardingLM, Lambert2024TLU3P, Ivison2023CamelsIA}. 
However, a key limitation of existing paradigms is that policy and reward models are usually developed in isolation, which restricts their interaction and reduces the potential for mutual improvement. 
In this work, we instead treat policy and reward as complementary capabilities, and introduce \methodname, a unified framework where the two evolve jointly, reinforcing each other and ultimately achieving stronger overall performance. We also discuss related work on self-reward and self-reflection; please refer to Appendix.~\ref{sec:appendix_related_works}.

\section{Methods}
In this section, we provide a detailed introduction to the \methodname approach. Specifically, Sec.~\ref{sec:preliminary} presents the \methodname training framework with verifiable reward, Sec.~\ref{sec:framework} outlines the on-policy reward\&reflection data generation of \methodname, and finally, Sec.~\ref{sec:tts} details the test-time scaling evaluation strategy used in  \methodname.

\subsection{\methodname Training with Verifiable Reward}\label{sec:preliminary}
\begin{figure}[t]
    \begin{center}
    \includegraphics[width=1.\linewidth]{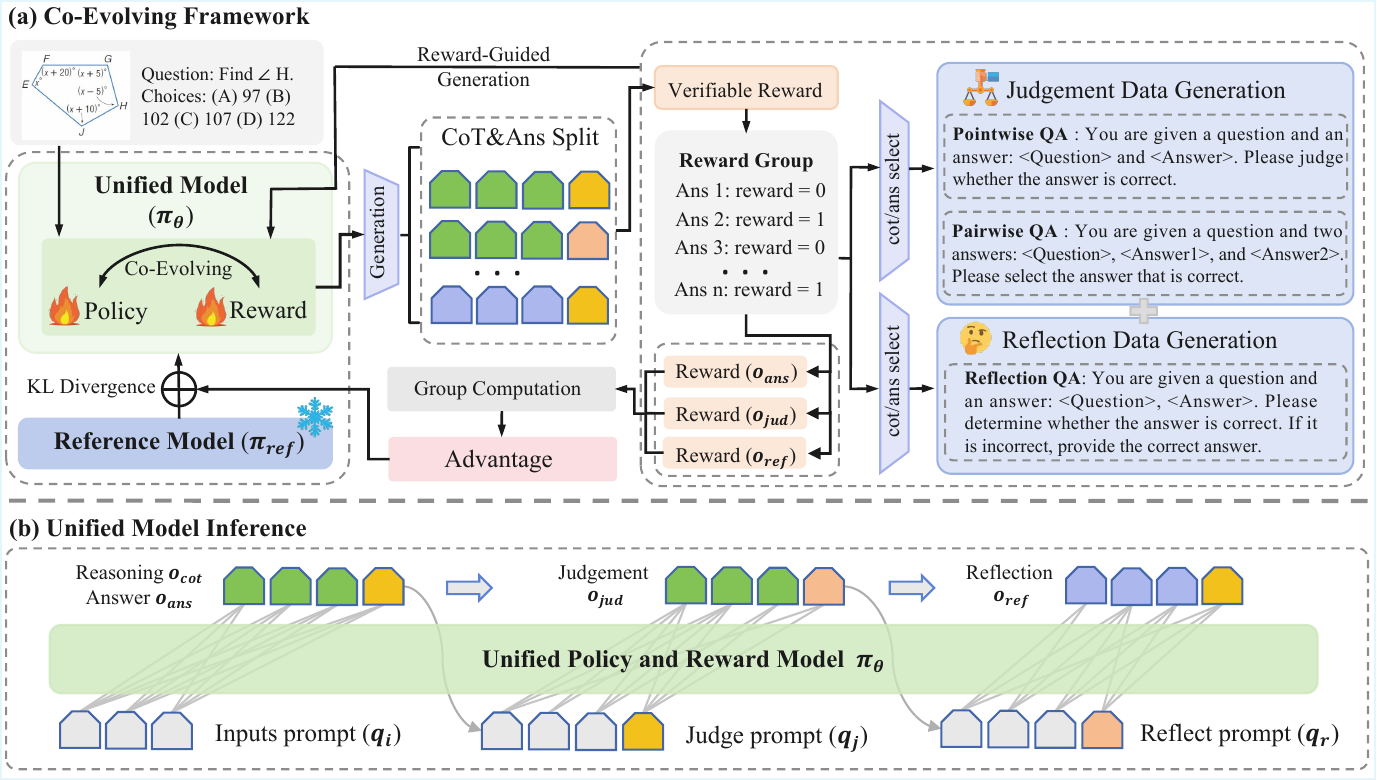}
    \end{center}
    \vspace{-12pt}
    \caption{\small \textbf{\methodname Framework.} 
    \textbf{(a) Training}: Our method recycles the valuable rollouts from verifiable reward-guided generation to simultaneously train a unified policy model $\pi_{\theta}$, also as a generative reward model. 
    \textbf{(b) Inference}: at test time, the single unified model can handle reasoning, judgment, and reflection tasks for test-time scaling, eliminating the need for external reward or judge models.
    }
    \vspace{-12pt}
    \label{fig:framework}
\end{figure}

Fig.~\ref{fig:framework} (a) illustrates our Synergistic Policy and Reward Co-Evolving Framework. In contrast to prior approaches, our method integrates the training of policy and reward into a unified framework, where both components are optimized within a single model under the guidance of verifiable rewards. In this section, we detail how \methodname employs verifiable rewards to guide optimization during training, enabling the model to co-evolve its policy and reward capabilities. Through this process, the model develops not only into a strong reasoning system but also into an effective reward model.

\paragraph{Step 1: Sampling an answer group.}
As shown in Fig.~\ref{fig:framework}(a), given a Visual Question Answering (VQA) sample $d=(q,a,I)$, or $d=(q,a)$ in the case of a language-only LLM without visual input, the model generates an answer group of size $n$, denoted as
\begin{equation}
G = \set*{ \bigl(o^{i}_{\text{cot}},\, o^{i}_{\text{ans}}\bigr) }_{i=1}^{n},
\end{equation}
where $q$ denotes the input prompt, $a$ the ground-truth answer, $I$ the input image, $o^{i}_{\text{cot}}$ the $i$-th reasoning trace, and $o^{i}_{\text{ans}}$ the corresponding final answer. To facilitate a clear separation between reasoning and the final output, we design a prompt that requires the answer to be enclosed in \texttt{\textbackslash box\{\}}, as illustrated in Appendix~\ref{sec:appendix_prompts}.
\paragraph{Step 2: Verifiable reward.}
Each final answer is evaluated by a rule-based, verifiable reward:
\begin{equation}
\mathbb{R}(q, o) =
\begin{dcases}
1, & \text{if } o = a, \\
0, & \text{otherwise}.
\end{dcases}
\label{eq:reward}
\end{equation}

For the $i$-th sample in the answer group $G$, we denote its reward as $r^{i}=\mathbb{R}\!\left(q,o^{i}_{\text{ans}}\right)$.

\paragraph{Step 3: Advantage computation.}
Following GRPO-style normalization, we compute a standardized advantage for each candidate:
\begin{equation}
\bar r=\frac{1}{n}\sum_{j=1}^{n} r^{j},\qquad
s=\sqrt{\frac{1}{n}\sum_{j=1}^{n}\bigl(r^{j}-\bar r\bigr)^{2}+\epsilon},\qquad
A^{i}=\frac{r^{i}-\bar r}{s},
\label{eq:advantage}
\end{equation}
where $\bar r$ is the mean reward across the $n$ candidates, $s$ is the standard deviation with a small constant $\epsilon>0$ for numerical stability, and $A^{i}$ is the normalized advantage used for policy gradient updates.

\paragraph{Step 4: Overall objective.}
The training objective maximizes the expected verifiable reward while regularizing the learned policy $\pi_{\theta}$ towards a reference policy $\pi_{\text{ref}}$:
\begin{equation}
\label{eq:rlvr_obj}
\mathbb{E}_{o\sim \pi_{\theta}(\cdot\,|\,q)}\big[\mathbb{R}(q,o)\big]
\;-\;
\lambda\,\mathrm{KL}\!\big(\pi_{\theta}(\cdot\,|\,q)\,\|\,\pi_{\text{ref}}(\cdot\,|\,q)\big),
\end{equation}
where $\lambda$ is a hyperparameter controlling the KL-divergence. This formulation ensures that the optimization signal comes directly from task-defined correctness, enabling efficient and stable training whenever outputs are objectively verifiable.

\subsection{\methodname On-Policy Data Generation}\label{sec:framework}
Unlike early RL methods that optimize only the policy model, the advantage of \methodname lies in the joint co-evolution of policy and reward within a single model. To achieve this, \methodname generates reward and reflection data on-policy during the reasoning process, which requires neither additional preference annotations nor teacher models, making it highly efficient.  

Beyond computing the advantage for gradient optimization in Eq.~\ref{eq:advantage}, the reward values $r^{i}$ also guide the on-policy generation of reward and reflection data, as illustrated in Fig.~\ref{fig:framework}. Specifically, we categorize the generated data into three forms: pointwise, pairwise, and reflection, each contributing to different aspects of judgment and self-reflection. 

\paragraph{Pointwise.}We reorganize $(q, G)$ into binary judgment samples of the form
\begin{equation}
\mathbb{D}_{\text{pointwise}}
= \big\{ \big(q,\; \boldsymbol{o}^{i}_{\text{ans}},\; \mathbb{R}(q, \boldsymbol{o}^{i}_{\text{ans}})\big) \big\},
\end{equation}

where the model is asked to determine whether a single candidate answer $o^{i}_{\text{ans}}$ is correct.  
This formulation directly trains the model’s ability to judge the validity of individual answers.  

\paragraph{Pairwise.}We also construct comparison-style samples:
\begin{equation}
\mathbb{D}_{\text{pairwise}}
= \Big\{ \big(q,\; \boldsymbol{o}^{i}_{\text{ans}},\; \boldsymbol{o}^{j}_{\text{ans}},\; 
\mathbb{R}(q,\boldsymbol{o}^{i}_{\text{ans}}),\; \mathbb{R}(q,\boldsymbol{o}^{j}_{\text{ans}})\big) \Big\},
\end{equation}

where two candidate answers $o^{i}_{\text{ans}}$ and $o^{j}_{\text{ans}}$ are drawn from $G$, and the model must select the better one.  
This encourages preference-style judgment, allowing the model to distinguish between relatively stronger and weaker outputs.  
Notably, in both pointwise and pairwise settings, $o_{\text{ans}}$ can be replaced with reasoning traces $o_{\text{cot}}$ to shift supervision toward intermediate steps.  

\paragraph{Reflection.}Finally, we construct reflection-style samples:
\begin{equation}
\mathbb{D}_{\text{reflect}}
= \Big\{ \big(q,\; \boldsymbol{o}^{i}_{\text{ans}},\; \mathbb{R}(q,\boldsymbol{o}^{i}_{\text{ans}})\big) \Big\},
\end{equation}
where the model first verifies correctness and, if $R(q, o^{i}_{\text{ans}})=0$, the incorrect answer is then fed back to the model for reflection and refinement.  
This process explicitly stimulates the model’s self-reflection capability.

The combined dataset is then given by  
\begin{equation}
\mathbb{D}_{\text{on-policy}}
= \mathbb{D}_{\text{pointwise}}
\;\cup\;
\mathbb{D}_{\text{pairwise}}
\;\cup\;
\mathbb{D}_{\text{reflect}},
\end{equation}
which is used to further optimize the unified policy–reward model, strengthening both its judgment and self-reflection abilities. Representative prompt templates used for data generation are provided in Appendix.~\ref{sec:appendix_prompts}.



\subsection{Test Time Scaling with Self-Reflection}\label{sec:tts} 

Benefiting from the co-evolution of policy and reward capabilities, \methodname functions not only as a strong policy model but also as a strong reward model. The synergy between these two abilities further enhances the model’s capacity for self-reflection, which proves especially valuable in the context of test-time scaling (TTS).  

As illustrated in Fig.~\ref{fig:framework}(b), we adopt a TTS procedure to evaluate the model’s reasoning, judgment, and reflection abilities. Formally, given a question $q$ and image $I$, the model generates a candidate answer at step $t$ as
\begin{equation}
    \mathbf{o}_t = \pi_{\boldsymbol{\theta}}(\, q, I \,), \label{eq:tts_gen}
\end{equation}
where $\pi_\theta$ denotes the model, and $o_t = (c_t, a_t)$ consists of a reasoning chain $c_t$ and a final prediction $a_t$. The model then assesses its own output through a judgment prompt:
\begin{equation}
\mathbf{r}_t = \pi_{\boldsymbol{\theta}}\!\bigl(q, I, \mathrm{judge}(c_t, a_t)\bigr),
\quad \mathbf{r}_t \in \{0,1\},
\end{equation}
where $\text{judge}(c_t, a_t)$ instructs the model to verify whether $(c_t, a_t)$ is correct.  

Based on this evaluation, the model either accepts the result or performs iterative refinement:
\begin{equation}
\mathbf{o}_{t+1} =
\begin{cases}
\mathbf{o}_t, & \text{if } \mathbf{r}_t = 1, \\[6pt]
\pi_{\boldsymbol{\theta}}\!\bigl(q, I, \mathrm{reflect}(c_t, a_t)\bigr), & \text{if } \mathbf{r}_t = 0,
\end{cases}
\end{equation}
where $\text{reflect}(c_t, a_t)$ prompts the model to critique its prior reasoning and generate a revised solution. The process terminates once the model produces an answer it judges correct, and accuracy is computed by comparing this final prediction with the ground truth.

\section{Experiments}

\subsection{Experimental Setup}

\paragraph{Benchmarks} 
To comprehensively evaluate the effectiveness of \methodname, we conduct experiments on three categories of benchmarks: mathematical, reward-related, and general.  
For mathematical benchmarks, we assess both multimodal and language-only reasoning using representative datasets such as MathVista~\citep{lu2023mathvista} and GSM8k~\citep{gsm8k}.  
For reward-related evaluation, we employ RewardBench2~\citep{malik2025rewardbench2} and VL-RewardBench~\citep{li2025vlrewardbench}, including their mathematical subsets for fine-grained analysis.  
For general capabilities, we test on widely used multimodal benchmarks such as MMBench~\citep{mmbench} and MMStar~\citep{mmstar}. For the complete list of benchmarks used in this work, please refer to Sec.~\ref{sec:benchmarks}.  

\paragraph{Baseline Methods} 
Our experiments are built upon the Qwen2.5-VL~\citep{bai2025qwen25vl} and Qwen2.5~\citep{yang2025qwen3} model series.  
For comparison, we include representative RL-based baselines such as VL-Rethinker~\citep{wang2025vlrethinker}, MM-Eureka~\citep{meng2025mmeureka}, Vision-R1~\citep{huang2025visionr1}, as well as the standard GRPO baselines, where \textit{Policy-Only} and \textit{Reward-Only} denote models trained to improve reasoning or judgment in isolation.  
Full details of the baseline models are provided in Sec.~\ref{sec:app_models} of the supplementary material.


\newcommand{\rot}[1]{\rotatebox[origin=c]{90}{\footnotesize #1}}

\begin{table}[t]
\caption{\small \textbf{Evaluation Results on \methodname-VL-7B.} We evaluate \methodname on multiple mathematical and reward-related benchmarks. Here, RB2 denotes RewardBench2, and VL-RB denotes VL-RewardBench.}
\label{tab:vlm7b_eval_math_reward}
\centering
\setlength{\tabcolsep}{4pt}         
\renewcommand{\arraystretch}{1.1}    
\resizebox{\linewidth}{!}{%
\begin{tabular}{@{} l *7{c} c *4{c} c c @{}}
\toprule
\multirow{2}{*}{\textbf{Model}} 
& \multicolumn{7}{c}{\textbf{VLM Math Benchmark}} & \cellcolor{gray!10}\textbf{Avg-M} 
& \multicolumn{4}{c}{\textbf{Reward Benchmark}} & \cellcolor{gray!10}\textbf{Avg-R} 
& \cellcolor{gray!20}\textbf{Avg-All} \\
\cmidrule(lr){2-8}\cmidrule(lr){10-13}
& \rot{MathVista} & \rot{MathVision} & \rot{WeMath} 
& \rot{MathVerse} & \rot{DynaMath} & \rot{LogicVista} & \rot{MMK12}
& \cellcolor{gray!10} 
& \rot{RB2} & \rot{RB2-Math} & \rot{VL-RB} & \rot{VL-RB-Math}
& \cellcolor{gray!10} 
& \cellcolor{gray!20} \\
\midrule
\addlinespace[1pt]
\textit{Baseline} \\[-2pt]
Qwen2.5-VL-7B  & 68.2 & 25.1 & 62.1 & 49.2 & 53.3 & 40.4 & 45.1 & \cellcolor{gray!10} 49.1 & 45.8 & 38.8 & 47.7 & 35.5 & \cellcolor{gray!10} 42.0 & \cellcolor{gray!20} 46.5\\
OpenVLThinker-7B & 70.2 & 25.3 & 64.3 & 47.9 & - & - & 60.6 & \cellcolor{gray!10} - & 48.5 & 37.3 & 33.2 & 33.1 & \cellcolor{gray!10} 38.0 & \cellcolor{gray!20} - \\
Vison-R1-7B & 73.5 & 27.4 & \textbf{75.0} & 52.4 & 54.9 & 37.1 & 36.7 & \cellcolor{gray!10} 51.0 & 32.6 & 28.1 & - & - & \cellcolor{gray!10} - & \cellcolor{gray!20} - \\
R1-OneVision-7B & 64.1 & \underline{29.9} & 61.8 &  47.1 & - & 39.1 & 39.8 & \cellcolor{gray!10} - & 35.7 & 33.1 & 37.6 & 37.4 & \cellcolor{gray!10} 36.0 & \cellcolor{gray!20} - \\
VL-ReThinker-7B & 73.7 & 28.4 & 67.9 & \textbf{54.0} & \underline{57.3} & 42.7 & 64.9 & \cellcolor{gray!10} 55.6 & 42.3 & 35.0 & 47.1 & 32.5 & \cellcolor{gray!10} 39.2 & \cellcolor{gray!20} 49.6\\
MM-Eureka-7B & 73.0 & 26.9 & 66.1 & 50.3 & 56.9 & \underline{48.9} & 64.5 & \cellcolor{gray!10} 55.2 & 44.9 & \underline{41.0} & 48.8 & 36.8 & \cellcolor{gray!10} 42.9 & \cellcolor{gray!20} 50.7\\
\midrule
\addlinespace[1pt]
\textit{Qwen2.5-VL-7B + GRPO} \\[-2pt]
$+$ Policy-Only   & 72.0 & 28.5 & 67.9 & 51.2 & 54.9 & 44.9 & 66.9 & \cellcolor{gray!10} 55.2 & 46.1 & 40.4 & 51.5 & 62.1 & \cellcolor{gray!10} 50.0 & \cellcolor{gray!20} 53.3\\
$+$ Reward-Only  & 72.1 & 27.9 & 67.1 & 51.0 & 54.7 & 44.0 & 58.8 & \cellcolor{gray!10} 53.7 & 48.1 & 39.3 & 53.9 & 62.7 & \cellcolor{gray!10} 51.0 & \cellcolor{gray!20} 52.7\\
$+$ Policy\&Reward & \underline{74.2} & 28.9 & \underline{70.9} & 51.3 & 56.3 & 46.2 & \underline{67.9} & \cellcolor{gray!10} \underline{56.5} & \underline{48.9} & \textbf{43.7} & \underline{54.4} & \underline{63.9} &  \cellcolor{gray!10} \underline{52.7} & \cellcolor{gray!20} \underline{55.1}\\
\midrule
\addlinespace[1pt]
\textit{Ours} \\[-2pt]
\rowcolor[HTML]{DAEFF9} \sparkvlsmall  & \textbf{75.9} & \textbf{31.1} & 70.3 & \underline{53.0} & \textbf{58.7} & \textbf{50.7} & \textbf{71.8} & \cellcolor{gray!10} \textbf{58.8} & \textbf{49.3} & 39.2 & \textbf{62.7} & \textbf{65.1} & \cellcolor{gray!10} \textbf{54.1} & \cellcolor{gray!20} \textbf{57.1}\\
$\Delta$  & \hgreen{+7.7} & \hgreen{+6.0} & \hgreen{+8.2} & \hgreen{+3.8} & \hgreen{+5.4} & \hgreen{+10.3} & \hgreen{+26.7} & \cellcolor{gray!10} \hgreen{+9.7} & \hgreen{+3.5} & \hgreen{+0.4} & \hgreen{+15.0} & \hgreen{+29.6} & \cellcolor{gray!10} \hgreen{+12.1} & \cellcolor{gray!20} \hgreen{+10.6} \\
\addlinespace[2pt]
\bottomrule
\end{tabular}%
}
\end{table}

\begin{figure}[t]
    \begin{center}
    \includegraphics[width=1.\linewidth]{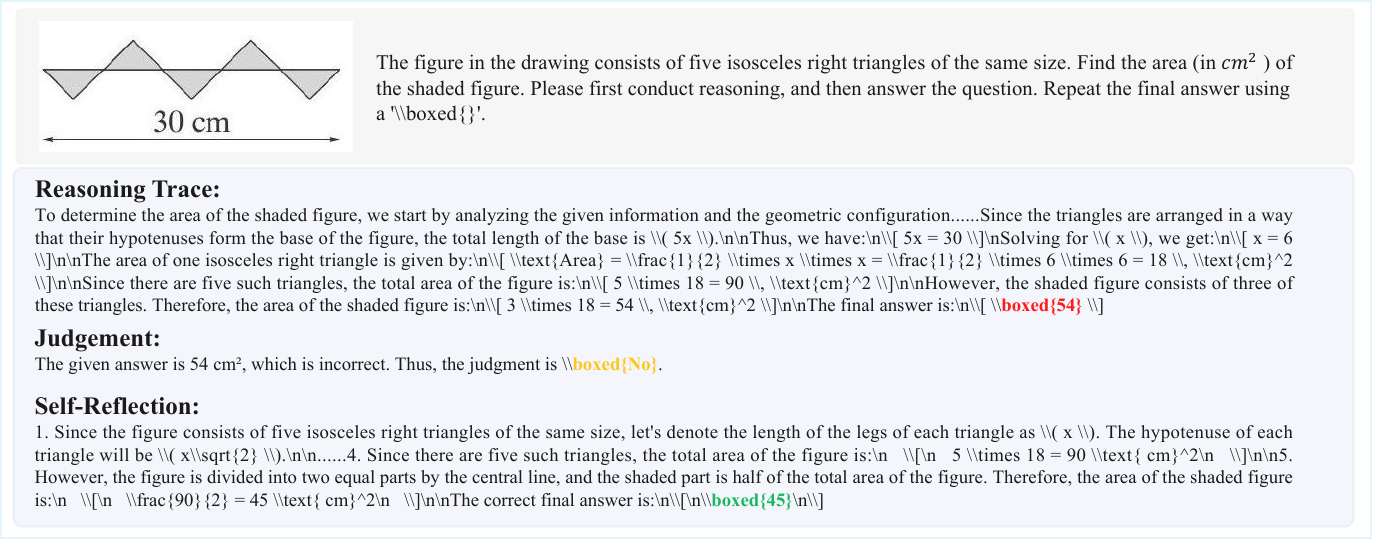}
    \end{center}
    \vspace{-18pt}
    \caption{\small \textbf{Math Reasoning Case.} We illustrate the reasoning process of \methodname on a mathematical task, covering reasoning, judgment, and reflection. For brevity, parts of the content are omitted.
    }
    \vspace{-12pt}
    \label{fig:math_case}
\end{figure}

\begin{table}[t]
\caption{\small \textbf{Evaluation Results on \methodname-7B.} We evaluate \methodname on multiple mathematical and reward-related benchmarks. Here, RB2 denotes RewardBench2.}
\label{tab:llm7b_eval_math_reward}
\centering
\setlength{\tabcolsep}{2pt}         
\renewcommand{\arraystretch}{1.1}    
\resizebox{\linewidth}{!}{%
\begin{tabular}{@{} l *6{c} c *2{c} c c @{}}
\toprule
\multirow{2}{*}{\textbf{Model}} 
& \multicolumn{6}{c}{\textbf{LLM Math Benchmark}} & \cellcolor{gray!10}\textbf{Avg-M} 
& \multicolumn{2}{c}{\textbf{Reward Benchmark}} & \cellcolor{gray!10}\textbf{Avg-R} 
& \cellcolor{gray!20}\textbf{Avg-All} \\
\cmidrule(lr){2-7}\cmidrule(lr){9-10}
& AIME24 & AIME25 & AMC23 
& GSM8k & Math-500 & MMLU-STEM
& \cellcolor{gray!10} 
& RB2 & RB2-Math
& \cellcolor{gray!10} 
& \cellcolor{gray!20} \\
\midrule
\addlinespace[1pt]
\textit{Baseline} \\[-2pt]
Qwen2.5-7B  & 6.7 & 6.7 & 50.0 & 91.9 & 76.2 & 75.8 & \cellcolor{gray!10} 51.2 & \underline{49.5} & 41.0 & \cellcolor{gray!10} 45.3 & \cellcolor{gray!20} 49.7\\
Simple-RL-Zero & 16.7 & 6.7 & \textbf{62.5} & 92.0 & 78.0 & 64.5 & \cellcolor{gray!10} \underline{53.4} & 31.0 & 38.7 & \cellcolor{gray!10} 34.9 & \cellcolor{gray!20} 48.8 \\
Eurus-2-7B-PRIME & \textbf{26.7} & - & \underline{57.8} & - & 79.2 & 71.0 & \cellcolor{gray!10} - & - &  - & \cellcolor{gray!10} - & \cellcolor{gray!20} - \\
Open-Reasoner-Zero-7B & 13.3 & - & 47.0 & - & 79.2 & 70.3 & \cellcolor{gray!10} - & 31.4 & 37.3 & \cellcolor{gray!10} 34.4 & \cellcolor{gray!20} - \\
Qwen2.5-Math & 13.3 & - & 50.6 & - & \textbf{79.8} & 60.2 & \cellcolor{gray!10} - & - & -  & \cellcolor{gray!10} - & \cellcolor{gray!20} - \\
\midrule
\addlinespace[1pt]
\textit{Qwen2.5-7B + GRPO} \\[-2pt]
$+$ Policy-Only & 6.7 & \underline{6.7} & 52.5 & 92.4 & 76.0 & 76.2 &\cellcolor{gray!10} 51.8 & 48.4 & 40.4 & \cellcolor{gray!10} 44.4 & \cellcolor{gray!20} 49.9\\
$+$ Reward-Only & 13.3 & 3.3 & 50.0 & 90.2 & 73.6 & 79.3  &\cellcolor{gray!10} 51.6 & 43.6 & 40.4 & \cellcolor{gray!10} 42.0 & \cellcolor{gray!20} 49.2\\
$+$ Policy\&Reward & 16.7 & 3.3 & 50.0 & \underline{92.7} & 76.6 & \underline{80.1} & \cellcolor{gray!10} 53.2 & 48.3 & \underline{43.7} & \cellcolor{gray!10} \underline{46.0} & \cellcolor{gray!20} \underline{51.4} \\
\midrule
\addlinespace[1pt]
\textit{Ours} \\[-2pt]
\rowcolor[HTML]{DAEFF9} \sparkllm & \underline{16.7} & \textbf{6.7} & \textbf{62.5} & \textbf{93.2} & \underline{79.4} & \textbf{81.1} & \cellcolor{gray!10} \textbf{56.6} & \textbf{58.8} & \textbf{55.7} & \cellcolor{gray!10} \textbf{57.3} & \cellcolor{gray!20} \textbf{56.8} \\
$\Delta$ & \hgreen{+10.0} & \hgreen{+0.0} & \hgreen{+12.5} & \hgreen{+1.3} & \hgreen{+3.2} & \hgreen{+5.3} &\cellcolor{gray!10} \hgreen{+5.4} & \hgreen{+9.3} & \hgreen{+14.7} & \cellcolor{gray!10} \hgreen{+12.0} & \cellcolor{gray!20} \hgreen{+7.1} \\
\addlinespace[2pt]
\bottomrule
\end{tabular}%
\vspace{-12pt}
}
\end{table}

\begin{table}[t]
\caption{\small \textbf{Evaluation Results on \methodname-VL-32B.} We evaluate \methodname on multiple mathematical and reward-related benchmarks. Here, RB2 denotes RewardBench2, and VL-RB denotes VL-RewardBench.}
\label{tab:vlm32b_reformat}
\centering
\setlength{\tabcolsep}{4pt}         
\renewcommand{\arraystretch}{1.1}    
\resizebox{\linewidth}{!}{%
\begin{tabular}{@{} l *7{c} c *4{c} c c @{}}
\toprule
\multirow{2}{*}{\textbf{Model}} 
& \multicolumn{7}{c}{\textbf{VLM Math Benchmark}} & \cellcolor{gray!10}\textbf{Avg-M} 
& \multicolumn{4}{c}{\textbf{Reward Benchmark}} & \cellcolor{gray!10}\textbf{Avg-R} 
& \cellcolor{gray!20}\textbf{Avg-All} \\
\cmidrule(lr){2-8}\cmidrule(lr){10-13}
& \rot{MathVista} & \rot{MathVision} & \rot{WeMath} 
& \rot{MathVerse} & \rot{DynaMath} & \rot{LogicVista} & \rot{MMK12}
& \cellcolor{gray!10} 
& \rot{RB2} & \rot{RB2-Math} & \rot{VL-RB} & \rot{VL-RB-Math}
& \cellcolor{gray!10} 
& \cellcolor{gray!20} \\
\midrule
\addlinespace[1pt]
\textit{Baseline} \\[-2pt]
Qwen2.5-VL-32B  & 74.7 & 38.4 & 69.1 & 48.5 & 61.3 & 55.4 & 52.9 & \cellcolor{gray!10} 57.2 & \underline{57.0} & 59.6 & \underline{59.2} & 56.0 & \cellcolor{gray!10} \underline{58.0} & \cellcolor{gray!20} 57.5 \\
VL-ReThinker-32B & \underline{78.8} & \textbf{40.5} & 76.7 & 56.9 & \textbf{62.9} & 51.8 & \underline{72.9} & \cellcolor{gray!10} 62.9 & 53.9 & 51.8 & 49.9 & 23.5 & \cellcolor{gray!10} 44.8 & \cellcolor{gray!20} 56.3 \\
Vision-R1-32B & 73.2 & 35.7 & \textbf{78.9} & 53.8 & \textbf{62.9} & 54.2 & 55.2 & \cellcolor{gray!10} 59.1 & 53.4 & \textbf{68.9} & - & - & \cellcolor{gray!10} - & \cellcolor{gray!20} - \\
MM-Eureka-32B & 74.8 & 34.4 & 73.4 & 56.5 & \underline{62.1} & 53.4 & 72.2 & \cellcolor{gray!10} 61.0 & 56.4 & 58.5 & 58.3 & \underline{56.6} & \cellcolor{gray!10} 57.5 & \cellcolor{gray!20} 59.7 \\
\midrule
\addlinespace[1pt]
\textit{Qwen2.5-VL-32B + GRPO} \\[-2pt] 	 	 	 	 	 	 	 		
 $+$ Policy\&Reward & 78.2 & \underline{40.2} & \underline{77.1} & \underline{57.7} & 60.9 & \underline{57.4} & 72.3 & \cellcolor{gray!10} \underline{63.4} & 56.3 & 56.4 & 57.1 & 53.6 & \cellcolor{gray!10} 55.9 & \cellcolor{gray!20} \underline{60.7}\\
\midrule
\addlinespace[1pt]
\textit{Ours} \\[-2pt]
\rowcolor[HTML]{DAEFF9} \sparkvlbig & \textbf{79.1} & \underline{40.2} & 76.7 & \textbf{59.2} & \textbf{62.9} & \textbf{59.4} & \textbf{77.4} & \cellcolor{gray!10} \textbf{65.0} & \textbf{60.3} & \underline{62.7} & \textbf{61.4} & \textbf{59.6} & \cellcolor{gray!10} \textbf{61.0} & \cellcolor{gray!20} \textbf{63.5}\\
$\Delta$  & \hgreen{+4.4} & \hgreen{+1.8} & \hgreen{+7.6} & \hgreen{+10.7} & \hgreen{+1.6} & \hgreen{+4.0} & \hgreen{+24.5} & \cellcolor{gray!10} \hgreen{+7.8} & \hgreen{+3.3} & \hgreen{+3.1} & \hgreen{+2.2} & \hgreen{+3.6} & \cellcolor{gray!10} \hgreen{+3.0} & \cellcolor{gray!20} \hgreen{+6.0} \\
\addlinespace[2pt]
\bottomrule
\end{tabular}%
\vspace{-12pt}
}
\end{table}

\begin{table}[t]
\caption{\small \textbf{Evaluation Results on General Multimodal Benchmarks.} We select multiple general multimodal benchmarks to assess the generalization and robustness of our method.}
\label{tab:vlm7b_eval_general_bench}
\centering
\setlength{\tabcolsep}{4pt}         
\renewcommand{\arraystretch}{1.15}  
\resizebox{\linewidth}{!}{%
\begin{tabular}{lccccccccc}
\toprule
\textbf{Models} & \textbf{MMBench} & \textbf{MMStar} & \textbf{MMMU} & \textbf{MMVet} & \textbf{ScienceQA} & \textbf{POPE} & \textbf{SeedBench} & \textbf{RealWorldQA} & \cellcolor{gray!20} \textbf{Average} \\
\midrule
Qwen2.5-VL-7B & 82.2 & 64.1 & 58.0 & \underline{69.7} & \underline{89.0} & 85.9 & \underline{77.0} & \underline{68.4} & \cellcolor{gray!20} 74.3 \\
VL-Rethinker-7B & 82.3 & \underline{65.4} & \textbf{59.0} & 69.3 & 87.8 & \underline{86.1} & 76.3 & 69.3 & \cellcolor{gray!20} \underline{74.4} \\
MM-Eureka-7B & \underline{84.2} & 65.3 & 57.8 & 68.9 & 88.8 & 85.8 & 76.8 & 65.1 & \cellcolor{gray!20} 74.1 \\
\midrule
\rowcolor[HTML]{DAEFF9} \sparkvlsmall & 
\textbf{84.4} & \textbf{67.3} & \underline{58.7} & \textbf{71.5} & \textbf{90.8} & \textbf{88.2} & \textbf{77.2} & \textbf{68.5} & \cellcolor{gray!20} \textbf{75.8} \\
$\Delta$ & 
\hgreen{+2.2} & \hgreen{+3.2} & \hgreen{+0.7} & \hgreen{+1.8} & \hgreen{+1.8} & \hgreen{+2.3} & \hgreen{+0.2} & \hgreen{+0.1} & \cellcolor{gray!20} \hgreen{+1.5} \\
\bottomrule
\end{tabular}%
\vspace{-12pt}
}
\end{table}

\begin{figure}[t]
    \begin{center}
    \includegraphics[width=1.\linewidth]{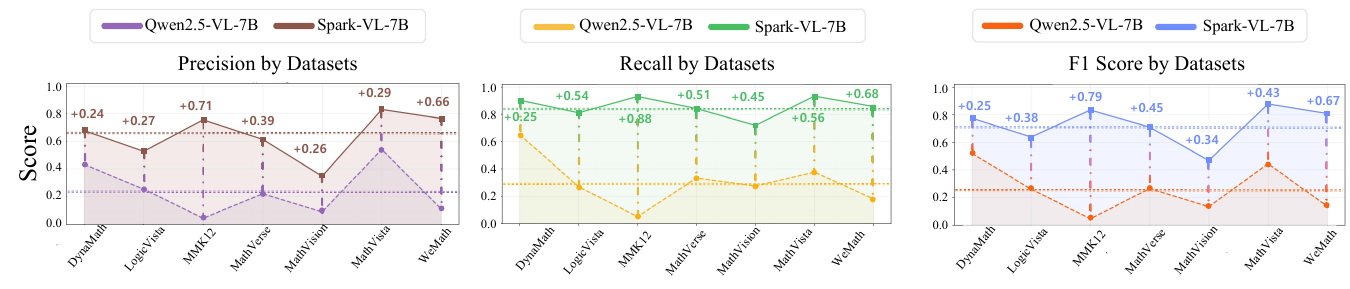}
    \end{center}
    \vspace{-12pt}
    \caption{\small \textbf{Study on Model’s Reward Accuracy.} We evaluate the model’s judgment ability by measuring its accuracy in determining whether its own answers are correct.
    }
    \vspace{-12pt}
    \label{fig:ablation_reward_accuracy}
\end{figure}

\subsection{Results on Mathematical and Reward Benchmarks}

\paragraph{Results on \sparkvlsmall.}
Tab.~\ref{tab:vlm7b_eval_math_reward} reports the results of \sparkvlsmall on both mathematical and reward-related benchmarks. Compared with the Qwen2.5-VL-7B baseline, our model achieves consistent and substantial gains. Specifically, \methodname delivers an average improvement of \hgreen{9.7\%} on mathematical benchmarks and \hgreen{12.1\%} on reward benchmarks, resulting in an overall gain of \hgreen{10.6\%}.

The \textit{Qwen2.5-VL-7B+GRPO} ablations further provide insight into the effect of different training signals. Training with \textit{Policy-Only} data slightly favors mathematical reasoning, while \textit{Reward-Only} training better enhances judgment ability. When both sources are combined (\textit{$+$Policy\&Reward}), the model surpasses either single-source variant, indicating the complementarity of policy and reward supervision. Building on this, \sparkvlsmall advances through the co-evolution of policy and reward capabilities, and further incorporates reflection-driven data generation, which strengthens the integration between reasoning and judgment and brings additional performance improvements.

Notably, from Tab.~\ref{tab:vlm7b_eval_math_reward} we also observe that \sparkvlsmall achieves significant improvements on both reward-related benchmarks, including RewardBench2 (\hgreen{+3.5\%}) and VL-RewardBench (\hgreen{+15.0\%}). Although all reward-related data generated during training are mathematics-specific, these two reward benchmarks span diverse domains. This indicates that \methodname is able to generalize its judgment ability beyond mathematics and perform strongly on broader tasks.

Overall, these results demonstrate that policy and reward are not in competition but instead mutually reinforcing. Joint optimization, when augmented with reflection, produces a synergistic effect that drives simultaneous and significant improvements in both reasoning and judgment. Representative reasoning cases on mathematical are illustrated in Fig.~\ref{fig:math_case}. More cases can be found in Appendix~\ref{sec:appendix_cases}.

\paragraph{Results on \sparkllm} 
To further assess the generalizability of our approach, we conduct additional experiments on the LLM Qwen2.5-7B. As shown in Tab.~\ref{tab:llm7b_eval_math_reward}, \sparkllm achieves average improvements of \hgreen{5.4\%} on mathematical benchmarks, \hgreen{12.0\%} on reward benchmarks, and \hgreen{7.1\%} overall, demonstrating consistent gains across diverse evaluation settings. 

Models such as Simple-RL-Zero~\citep{zeng2025simplerl} and the \textit{Policy-Only} variant of Qwen2.5-7B show declines on reward benchmarks, which reflects the trade-off between reasoning and judgment ability. Moreover, we observe that \textit{Reward-Only} GRPO training performs worse than \textit{Policy-Only} training on both reasoning and reward tasks. This suggests that the model tends to overfit to reward signals, which in turn weakens its reasoning ability and prevents it from excelling in either skill. The \textit{Policy\&reward} GRPO variant partially alleviates this issue by training both capabilities simultaneously. Building on this, \sparkllm further leverages on-policy co-evolution and reflection mechanisms to more effectively fuse and reinforce reasoning and judgment, ultimately achieving a substantial performance leap.

\paragraph{Results on Qwen2.5-VL-32B}
To further examine the scalability of our approach, we conduct experiments with Qwen2.5-VL-32B as the backbone. As shown in Tab.~\ref{tab:vlm32b_reformat}, \sparkvlbig achieves improvements of \hgreen{+7.8\%} on mathematical benchmarks and \hgreen{+3.0\%} on reward benchmarks. These results confirm that \methodname scales effectively to larger models, with the co-evolving mechanism continuing to deliver consistent gains by leveraging the richer capacity of the backbone. This highlights both the robustness of our framework and its potential for application to even stronger models.

\subsection{Results on General Benchmarks}
We further assess our approach on a suite of general-purpose benchmarks to evaluate its capabilities beyond mathematics and judgment tasks. As presented in Tab.~\ref{tab:vlm7b_eval_general_bench}, \sparkvlsmall achieves improvements of \hgreen{2.2\%}, \hgreen{3.2\%}, and \hgreen{1.8\%} on MMBench~\citep{mmbench}, MMStar~\citep{mmstar}, and MMVet~\citep{mmvet}, respectively, leading to an average gain of \hgreen{1.5\%} across eight benchmarks. Compared with other reinforcement learning approaches, \methodname extends its reasoning and reflection abilities beyond the mathematical domain, demonstrating stronger generalization to diverse tasks.

\subsection{Judgment Accuracy Analysis}\label{sec:reward_analysis}
To further compare the judgment ability of \methodname with the Qwen2.5-VL-7B baseline, we conduct an additional evaluation on self-judgment accuracy. Specifically, we use data from seven mathematical datasets as input: the model is required to first perform reasoning and then assess the correctness of its own prediction. Based on these judgments, we compute recall, precision, and F1 scores. The results, shown in Fig.~\ref{fig:ablation_reward_accuracy}, indicate that \methodname consistently outperforms the baseline across all three metrics, with particularly pronounced gains on MMK12~\citep{meng2025mmeureka}, MathVerse~\citep{zhang2024mathverse}, and WeMath~\citep{qiao2024wemath}. These findings demonstrate that our training framework enables strong self-judgment ability without relying on manually annotated preference data or larger teacher models.

\subsection{Ablation Studies}\label{sec:ablation_study}

\paragraph{Ablation Study on Test-Time Scaling}
In \methodname, we adopt a reflection-augmented test-time scaling (TTS) strategy to activate the model’s integrated capabilities of reasoning, judgment, and self-reflection. As shown in Tab.~\ref{tab:vlm7b_ablation_reflection}, applying TTS to Qwen2.5-VL-7B leads to a noticeable performance drop across multiple benchmarks, primarily because its weak judgment and reflection skills cause frequent misjudgments, especially when the number of reasoning rounds increases. For the \textit{GRPO$+$Policy} setting, TTS yields only marginal improvements. In contrast, \methodname achieves substantially larger gains, benefiting from its inherently enhanced reasoning, judgment, and reflection capabilities.

\begin{table}[t]
\caption{\small \textbf{Ablation on Test-Time Scaling.} We conduct ablation studies on the TTS. Specifically, we apply judge-reflection-based TTS to both the Qwen2.5-VL-7B model and the GRPO-trained model to evaluate its effectiveness.}
\label{tab:vlm7b_ablation_reflection}
\centering
\setlength{\tabcolsep}{2pt}         
\renewcommand{\arraystretch}{1.0}    
\resizebox{\linewidth}{!}{%
\begin{tabular}{@{} l *7{c} c @{}}
\toprule
\textbf{Model}
& MathVista & MathVision & WeMath 
& MathVerse & DynaMath & LogicVista & MMK12
& \cellcolor{gray!20} Average\\
\midrule
\addlinespace[1pt]
\textit{Baseline} \\[-2pt]
Qwen2.5-VL-7B  & 68.2 & 25.1 & 62.1 & 49.2 & 53.3 & 40.4 & 45.1 & \cellcolor{gray!20} 49.1\\
Qwen2.5-VL-7B+TTS  & 68.5 & 18.4 & 24.1 & 29.9 & 52.7 & 43.1 & 42.8 & \cellcolor{gray!20} 39.9\\
\midrule
\addlinespace[1pt]
\textit{Qwen2.5-VL-7B + GRPO} \\[-2pt]
 $+$ Policy   & 72.0 & 28.5 & 67.9 & 51.2 & 54.9 & 44.9 & 66.9 & \cellcolor{gray!20} 55.2\\
 $+$ Policy \& TTS  & 73.1 & 28.7 & 67.5 & 51.9 & 57.7 & 48.9 & 68.9 & \cellcolor{gray!20} 56.6 \\
\midrule
\rowcolor[HTML]{DAEFF9} \sparkvlsmall  & 75.9 & 31.1 & 70.3 & 53.0 & 58.7 & 50.7 & 71.8 & \cellcolor{gray!20} 58.8 \\
\addlinespace[2pt]
\bottomrule
\end{tabular}%
}
\end{table}

\begin{table}[t]
\caption{\small \textbf{Ablation Study on Answer- vs. CoT-based Data Generation.} We generate data on-policy using either final answers, chains of thought (CoT), or a combination of both, and evaluate the impact of these strategies on performance.}
\label{tab:vlm7b_ablation_ans_cot}
\centering
\setlength{\tabcolsep}{2pt}         
\renewcommand{\arraystretch}{1.1}    
\resizebox{\linewidth}{!}{%
\begin{tabular}{@{} l *7{c} c *4{c} c c @{}}
\toprule
\multirow{2}{*}{\textbf{Model}} 
& \multicolumn{7}{c}{\textbf{VLM Math Benchmark}} & \cellcolor{gray!10}\textbf{Avg-M} 
& \multicolumn{4}{c}{\textbf{Reward Benchmark}} & \cellcolor{gray!10}\textbf{Avg-R} 
& \cellcolor{gray!20}\textbf{Avg-All} \\
\cmidrule(lr){2-8}\cmidrule(lr){10-13}
& MathVista & MathVision & WeMath 
& MathVerse & DynaMath & LogicVista & MMK12
& \cellcolor{gray!10} 
& RB2 & RB2-Math & VL-RB & VL-RB-Math
& \cellcolor{gray!10} 
& \cellcolor{gray!20} \\
\midrule
\methodname + Ans  & 73.8 & 29.1 & 69.0 & 51.9 & 58.1 & 46.2 & 69.9 & \cellcolor{gray!10} 56.9 & 47.2 & 41.8 & 57.9 & 63.9 & \cellcolor{gray!10} 52.7 & \cellcolor{gray!20} 55.3\\
\methodname + CoT  & 73.6 & 29.7 & 67.4 & 50.4 & 57.3 & 48.4 
& 71.3 & \cellcolor{gray!10} 56.9 & 52.5 & 44.0 & 62.3 & 60.8 & \cellcolor{gray!10} 54.9 & \cellcolor{gray!20} 56.2\\
\rowcolor[HTML]{DAEFF9} \methodname + Ans\&CoT  & 75.9 & 31.1 & 70.3 & 53.0 & 58.7 & 50.7 & 71.8 &  58.8 & 49.3 & 39.2 & 62.7 & 65.1 &  54.1 &  57.1\\
\addlinespace[2pt]
\bottomrule
\end{tabular}%
\vspace{-12pt}
}
\end{table}

\paragraph{Ablation Study on Answer- vs. CoT-based Data Generation}  
In \methodname training, on-policy reward data can be generated either from final answers or from chains of thought (CoT). As shown in Tab.~\ref{tab:vlm7b_ablation_ans_cot}, we compare three settings: using only answer-based data (55.3), using only CoT-based data (56.2), and combining both (57.1). The results indicate that integrating both sources leads to the best performance, suggesting that the complementary nature of answer- and CoT-based data provides richer training signals and ultimately enhances the model’s learning effectiveness.

\paragraph{Cost Analysis}\label{sec:cost_analysis}
\begin{wraptable}{r}{0.50\linewidth} 
\centering
\caption{\small \textbf{Comparison between RM-based RL and \methodname.}}
\scalebox{0.65}{ 
\begin{tabular}{lcc}
\toprule
 & \textbf{RM-based RL} & \textbf{\methodname (Ours)} \\
\midrule
\textbf{Extra Data (Preference)} & \ding{51}  & \ding{55} \\
\textbf{Extra RM Training } & \ding{51} & \ding{55} \\
\textbf{GPU Cost} & $\sim$2$\times$ & 1$\times$  \\
\textbf{Reward Signal} & RM inference & Rule-based signal \\
\textbf{Efficiency} & Slower & Faster \\
\bottomrule
\end{tabular}
}
\label{tab:rm_vs_spark}
\end{wraptable}
Unlike traditional RM-based RL methods, \methodname removes the need for an additional reward model and extra preference data. As shown in Tab.~\ref{tab:rm_vs_spark}, RM-based RL requires a separate RM training stage with large-scale human or teacher annotations, and during RL optimization, it repeatedly calls the RM for reward inference, which doubles GPU usage and slows training. In contrast, \methodname directly employs lightweight rule-based verifiable rewards to generate feedback on-policy, allowing a single unified model to optimize both policy and reward. This design not only reduces data and computational costs but also ensures a faster and more scalable training pipeline.

\section{Conclusion}

We presented the \textbf{S}ynergistic \textbf{P}olicy \textbf{A}nd \textbf{R}eward Co-Evolving Framewor\textbf{K} (\textbf{\methodname}), an efficient, on-policy, and stable paradigm that unifies policy optimization and reward modeling within a single model. Unlike prior RL pipelines that treat policy and reward in isolation or rely on costly external reward models, \methodname recycles RLVR rollouts into judgment and reflection objectives, enabling the model itself to function as both a strong policy and a generative reward model. This co-evolving mechanism establishes a positive feedback loop: improved reward accuracy enhances reasoning ability, while stronger reasoning in turn refines reward judgment, fostering self-reflection and stability. Demonstrating substantial improvements on mathematical, reward, and general benchmarks, \methodname offers a scalable and generalizable solution for RL, advancing a new paradigm where reasoning, judgment, and reflection evolve synergistically.

\bibliography{iclr2026_conference}
\bibliographystyle{iclr2026_conference}

\newpage
\appendix
\section{Appendix}


\section*{Outline}
In the appendix, we provide additional supporting materials to facilitate a deeper understanding of our work.  
First, in Sec.~\ref{sec:appendix_dataset_benchmark_statistic}, we summarize all the models, datasets, and benchmarks used in the experiments of \methodname.  
Second, in Sec.~\ref{sec:appendix_prompts}, we present the prompt templates employed in our study, including those used during evaluation as well as the on-policy prompt designs adopted by \methodname for reward and reflection data generation.  
Third, Sec.~\ref{sec:appendix_related_works} discusses related works on self-reward and self-reflection. 
Finally, in Sec.~\ref{sec:appendix_cases}, we provide several illustrative reasoning cases from \methodname on mathematical and reward benchmarks.

\subsection{Model, Dataset and Benchmark Statistic}\label{sec:appendix_dataset_benchmark_statistic}

\subsubsection{Models}\label{sec:app_models}
In our study, we adopt the Qwen family of models as the backbone, including Qwen2.5-VL-7B~\citep{bai2025qwen25vl}, Qwen2.5-VL-32B~\citep{bai2025qwen25vl}, and Qwen2.5-7B~\citep{yang2025qwen3}. Based on these backbones, we train three corresponding variants of our proposed framework: \sparkvlsmall, \sparkvlbig, and \sparkllm. These variants allow us to comprehensively evaluate the effectiveness of \methodname across both multimodal and text-only settings, as well as across different model scales.  

For comparison, we benchmark against a wide range of existing RL-based approaches. In the multimodal domain, we include VL-Rethinker-7B~\citep{wang2025vlrethinker}, MM-Eureka-7B~\citep{meng2025mmeureka}, OpenVLThinker-7B~\citep{deng2025openvlthinker}, Vision-R1-7B~\citep{huang2025visionr1}, and R1-OneVision-7B~\citep{yang2025r1onevision}, all of which represent recent efforts to strengthen reasoning capacity in vision–language models through reinforcement learning. In the language domain, we compare with Qwen2.5-Math-7B-Instruct~\citep{qwen25math}, Simple-RL-Zero-7B~\citep{zeng2025simplerl}, Eurus-2-7B-PRIME~\citep{cui2025eurus-prime}, and Open-Reasoner-Zero-7B~\citep{hu2025openreasonerzero}, which focus primarily on mathematical or general reasoning tasks within the NLP setting.  

\begin{table}[h]
\caption{
\small
\textbf{Model Sources.} We have compiled a list of all the models involved in the experiments along with their parameter scale.}
\label{tab:model_card}
\begin{center}
\setlength{\tabcolsep}{2pt}
\scalebox{.90}{
\begin{tabular}{ll}
\toprule
\multicolumn{1}{c}{\bf Models} &\multicolumn{1}{c}{\bf Parameter}\\ 
\midrule
\addlinespace[1pt]
\textit{Baseline} \\[-2pt]
Qwen2.5-VL-7B-Instruct~\citep{bai2025qwen25vl} & 7B \\
Qwen2.5-VL-32B-Instruct~\citep{bai2025qwen25vl} & 32B  \\
Qwen2.5-7B-Instruct~\citep{yang2025qwen3} & 7B \\
\midrule
\addlinespace[1pt]
\textit{Multimodal} \\[-2pt]
VL-Rethinker-7B~\citep{wang2025vlrethinker}& 7B \\
VL-Rethinker-32B~\citep{wang2025vlrethinker}& 32B \\
MM-Eureka-7B~\citep{meng2025mmeureka}& 7B\\
MM-Eureka-32B~\citep{meng2025mmeureka}& 32B  \\
OpenVLThinker-7B~\citep{deng2025openvlthinker}& 7B \\
Vision-R1-7B~\citep{huang2025visionr1}& 7B  \\
Vision-R1-32B~\citep{huang2025visionr1}& 32B  \\
R1-OneVision-7B~\citep{yang2025r1onevision}& 7B \\
\midrule
\addlinespace[1pt]
\textit{Language-Only} \\[-2pt]
Qwen2.5-Math-7B-Instruct~\citep{qwen25math} & 7B  \\
Simple-RL-Zero-7B~\citep{zeng2025simplerl} & 7B  \\
Eurus-2-7B-PRIME~\citep{cui2025eurus-prime} & 7B \\
Open-Reasoner-Zero-7B~\citep{hu2025openreasonerzero} & 7B \\
\bottomrule
\end{tabular}
\vspace{-12pt}
}
\end{center}
\end{table}

To further examine the generalization ability of \methodname across different scales, we additionally evaluate against larger multimodal baselines, including VL-Rethinker-32B~\citep{wang2025vlrethinker}, MM-Eureka-32B~\citep{meng2025mmeureka}, and Vision-R1-32B~\citep{huang2025visionr1}. These larger-scale models provide an important reference point to test whether the improvements introduced by \methodname are preserved when scaling up.  

A complete summary of all models used in our experiments, along with their categories (multimodal vs. language-only, 7B vs. 32B scale), is provided in Tab.~\ref{tab:model_card} of the supplementary material.



\begin{table}[h]
\caption{
\small
\textbf{Benchmark Sources.} We have included information for all the benchmarks tested in the paper in the table.}
\label{tab:benchmark_source}
\begin{center}
\setlength{\tabcolsep}{4pt}
\scalebox{.9}{
\begin{tabular}{l|l}
\toprule
\multicolumn{1}{c}{\bf Setting} &\multicolumn{1}{c}{\bf Models} \\ 
\midrule
\multirow{7}{*}{\begin{tabular}[c]{@{}c@{}} \bf Mathematical\\ \bf Multimodal \\ \bf Benchmark
\end{tabular}} & MathVista~\citep{lu2023mathvista} \\
~ & MathVision~\citep{mathvision} \\
~ & WeMath~\citep{qiao2024wemath} \\
~ & MathVerse~\citep{zhang2024mathverse} \\
~ & DynaMath~\citep{zou2024dynamath} \\
~ & LogicVista~\citep{xiao2024logicvista}  \\
~ & MMK12~\citep{meng2025mmeureka} \\
\midrule
\multirow{2}{*}{\begin{tabular}[c]{@{}c@{}} \bf Reward\\ \bf Benchmark 
\end{tabular}} & RewardBench2~\citep{malik2025rewardbench2}\\
~ & VL-RewardBench~\citep{li2025vlrewardbench} \\
\midrule
\multirow{8}{*}{\begin{tabular}[c]{@{}c@{}} \bf General\\ \bf Multimodal\\ \bf Benchmark 
\end{tabular}} & MMMU~\citep{yue2024mmmu} \\
~ & MMVet~\citep{mmvet} \\
~ & MMBench~\citep{mmbench} \\
~ & MMStar~\citep{mmstar} \\
~ & POPE~\citep{pope} \\
~ & ScienceQA~\citep{scienceqa} \\
~ & SeedBench~\citep{ying2025seedbench} \\
~ & RealWorldQA \\
\bottomrule
\end{tabular}
\vspace{-12pt}
}
\end{center}
\end{table}

\subsubsection{Benchmarks} \label{sec:benchmarks}
We evaluate \methodname across three major categories of benchmarks: mathematical reasoning, reward-related evaluation, and general multimodal understanding. This comprehensive setup ensures that our analysis covers not only specialized domains but also broader tasks.  

Mathematical Benchmarks. For multimodal mathematical reasoning, we adopt MathVista~\citep{lu2023mathvista}, MathVision~\citep{mathvision}, WeMath~\citep{qiao2024wemath}, MathVerse~\citep{zhang2024mathverse}, DynaMath~\citep{zou2024dynamath}, LogicVista~\citep{xiao2024logicvista}, and MMK12~\citep{meng2025mmeureka}.  
For text-only reasoning, we include AIME24, AIME25, AMC23, GSM8k~\citep{gsm8k}, Math500~\citep{math500}, and the MMLU\_STEM~\citep{mmlu}. These datasets collectively test numerical reasoning, symbolic manipulation, logical deduction, and competition-style problem solving under both textual and multimodal settings.  

Reward-related Benchmarks. We adopt RewardBench2 (RB2)~\citep{malik2025rewardbench2} and VL-RewardBench (VL-RB)~\citep{li2025vlrewardbench} as representative reward evaluation benchmarks. Both focus on assessing models’ ability to judge correctness and quality of generated outputs. In addition, we separately report results on the mathematical subsets of these benchmarks, in order to analyze the interplay between reward judgment and mathematical reasoning.  

General-purpose Multimodal Benchmarks. To assess generalization and robustness beyond math and reward domains, we evaluate on a diverse set of multimodal benchmarks, including MMMU~\citep{yue2024mmmu}, MMVet~\citep{mmvet}, MMBench~\citep{mmbench}, MMStar~\citep{mmstar}, POPE~\citep{pope}, ScienceQA~\citep{scienceqa}, SeedBench~\citep{ying2025seedbench}, and RealWorldQA. These benchmarks cover a wide spectrum of multimodal reasoning and understanding tasks, ranging from knowledge-intensive QA to real-world perception challenges.  

A complete statistical summary of all benchmarks is provided in Tab.~\ref{tab:benchmark_source}.

\subsubsection{Training Data Preparation}\label{sec:training_data_pre} 
Previous approaches that aimed to train a model’s judgment or reflection capabilities typically relied on either reward data generated by a teacher model or manually annotated reflection traces. In contrast, \methodname is able to generate the required reward and reflection training data on-policy during policy’s RFT, guided by verifiable reward signals. This approach not only greatly reduces the cost of data collection, but also ensures that the generated data evolves alongside model optimization and remains consistently aligned with the model’s current policy distribution.

Leveraging \methodname’s ability to generate reward and reflection data on-policy, we only need to collect VQA triples consisting of images, questions, and answers. In our experiments, 19k randomly sampled instances from ViRL-39k~\citep{wang2025vlrethinker} are used to train \sparkvlsmall, while 24k difficulty-filtered instances from the same dataset are used for \sparkvlbig. For the language-only variant \sparkllm, we employ the Simple-RL-Zero-25k dataset~\citep{zeng2025simplerl}. All data are represented in the form of $(q, a, I)$, where $q$ denotes the question, $a$ the ground-truth answer, and $I$ the corresponding image. Notably, this setup requires no manually annotated reward data, reflection traces, or judgment-oriented CoT trajectories. 

\begin{figure}[t]
    \begin{center}
    \includegraphics[width=1.\linewidth]{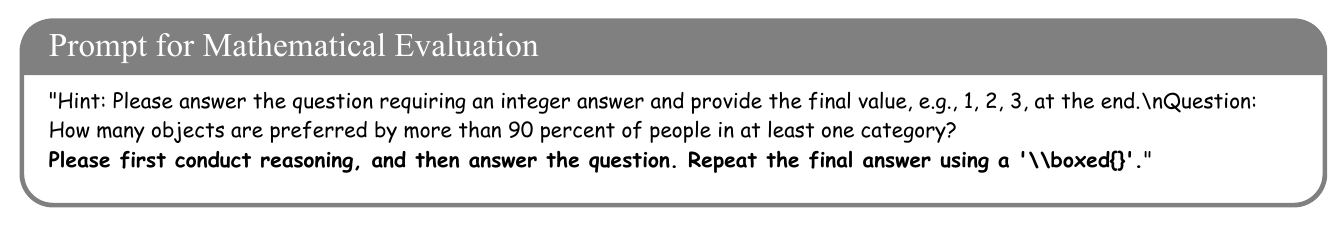}
    \end{center}
    \caption{\small \textbf{Mathematical Prompt.} Prompt suffix used for mathematical benchmark evaluation.}
    \vspace{-12pt}
    \label{fig:math_prompt_example}
\end{figure}

\begin{figure}[t]
    \begin{center}
    \includegraphics[width=1.\linewidth]{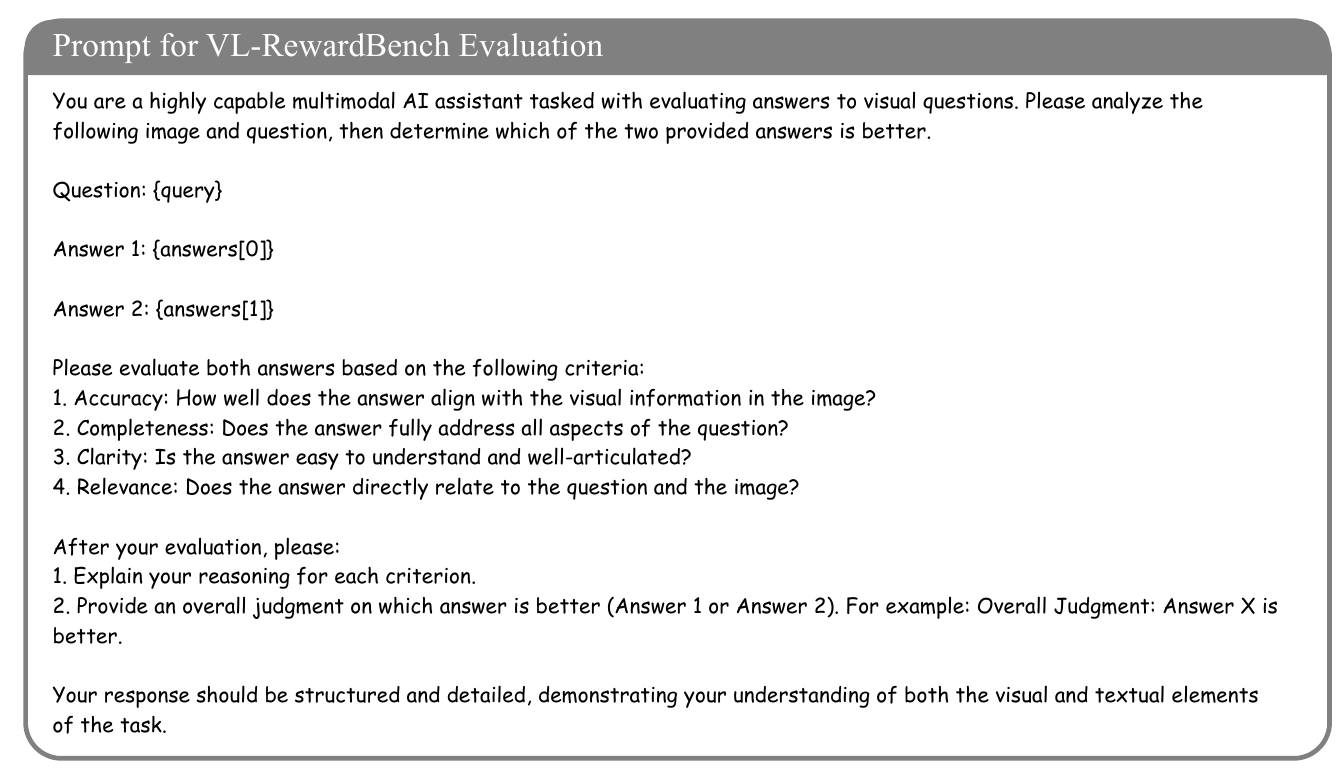}
    \end{center}
    \caption{\small \textbf{Prompt for VL-RewardBench.} In the figure, the placeholders \textit{query} and \textit{answer} should be replaced with the specific content of each task.}
    \vspace{-12pt}
    \label{fig:reward_prompt_example}
\end{figure}

\begin{figure}[t]
    \begin{center}
    \includegraphics[width=1.\linewidth]{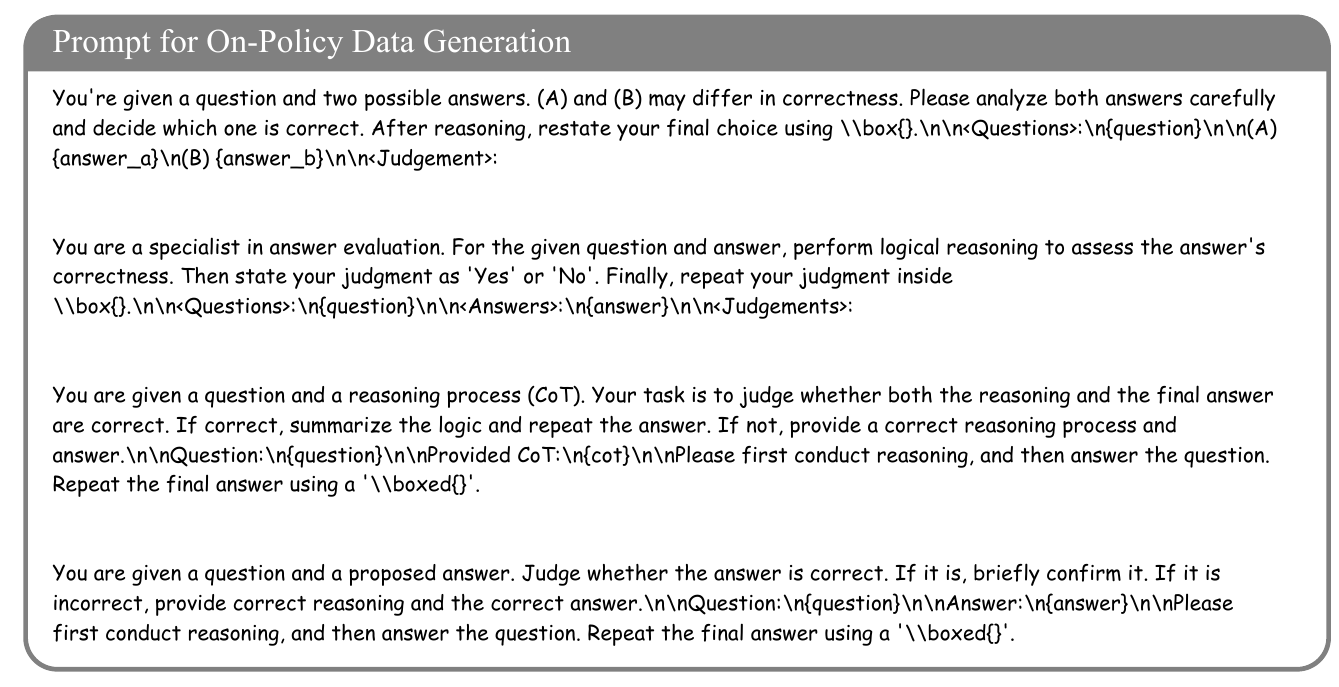}
    \end{center}
    \caption{\small \textbf{On-Policy Data Generation Prompts.} The figure illustrates four different prompt templates used for generating reward and reflection data.}
    \vspace{-12pt}
    \label{fig:data_gen_prompt_example}
\end{figure}

\subsection{Prompts}\label{sec:appendix_prompts}

\paragraph{Evaluation Prompts} During dataset evaluation, we appended an additional prompt at the end of each mathematical problem to facilitate the separation of reasoning steps from the final answer. To avoid over-constraining the model with rigid output formats (e.g., \texttt{<think><answer>}), we instead instructed the model to enclose the final answer within \texttt{\textbackslash box\{\}} after completing its reasoning process. A concrete example is provided in Fig.~\ref{fig:math_prompt_example}.

For reward-related benchmarks, we followed the official evaluation prompts provided by each benchmark. For instance, in VL-RewardBench~\citep{li2025vlrewardbench}, we adopted the original prompt format as illustrated in Fig.~\ref{fig:reward_prompt_example}.

\paragraph{On-Policy Data Generation Prompts} A key step in the policy–reward co-evolving training of \methodname is the on-policy generation of reward and reflection data. During GRPO optimization, the reward signals serve two purposes: on the one hand, they are used to compute the advantage for updating model parameters; on the other hand, they guide the construction of reward and reflection data. Specifically, chain of thought (CoT) or final answers filtered by reward values are wrapped with carefully designed prompts and reorganized into new training samples. These samples further enhance the model’s judgment and reflection abilities. Examples of the prompts used for reward and reflection data generation are shown in Fig.~\ref{fig:data_gen_prompt_example}.

\subsection{Related Works}\label{sec:appendix_related_works}
\paragraph{Self-Reward and Self-Reflection.}
Early studies have explored incorporating \textit{self-reward} and \textit{self-reflection} into supervised fine-tuning (SFT) pipelines by generating reward or reflection data to enhance reasoning abilities. For example, STaR~\citep{Zelikman2022STaRBR} iteratively generates chain-of-thought traces to improve its own reasoning capability. S2R~\citep{Ma2025S2RTL} employs pre-annotated self-verification and self-correction data for both SFT and RL training. COOPER~\citep{Hong2025CooperCP} leverages an external assistant to generate preference data, which are then used to train a reward model. While these approaches demonstrate the potential of self-reward and self-reflection for improving reasoning, they still rely on either external annotation data or independently trained reward models. In contrast, our method is the first to unify policy and reward capabilities within a single model by optimizing the GRPO framework. This co-evolving design breaks the conventional paradigm of separately trained reward models, enabling policy and reward to mutually reinforce each other, and integrates reasoning, judgment, and self-reflection into a unified process—without the need for preference annotation or external reward modeling.

\subsection{Case Study}\label{sec:appendix_cases}
\begin{figure}[t]
    \begin{center}
    \includegraphics[width=1.\linewidth]{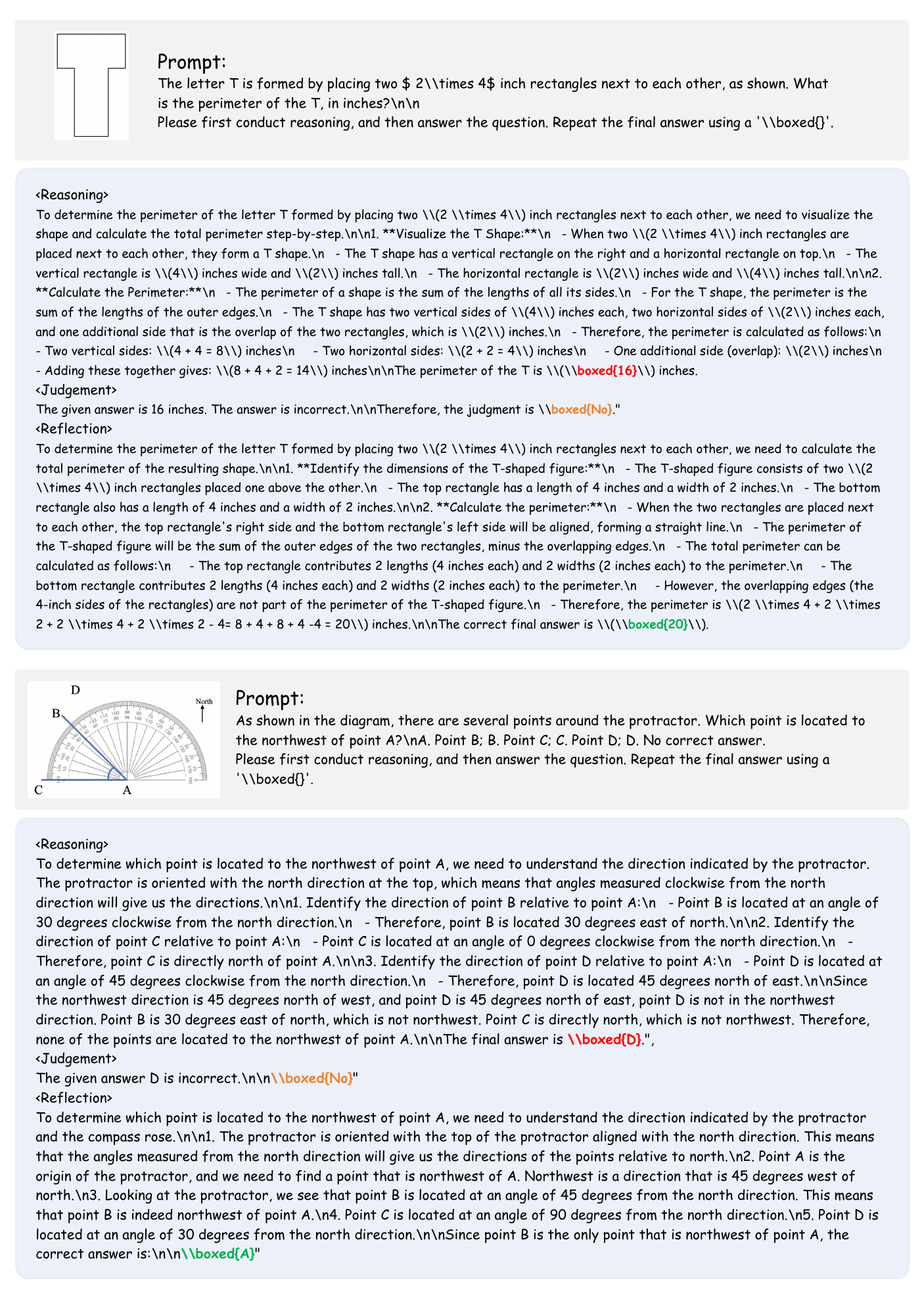}
    \end{center}
    \caption{\small \textbf{Math Inference Case.}}
    \vspace{-12pt}
    \label{fig:math_case_sup}
\end{figure}

In Fig.~\ref{fig:math_case_sup}, we present several reasoning cases on mathematical problems, which provide an intuitive demonstration of \methodname’s integrated capabilities in reasoning, judgment, and reflection. The selected examples highlight scenarios where the model engages in both reasoning and self-judgment.  

\begin{figure}[t]
    \begin{center}
    \includegraphics[width=1.\linewidth]{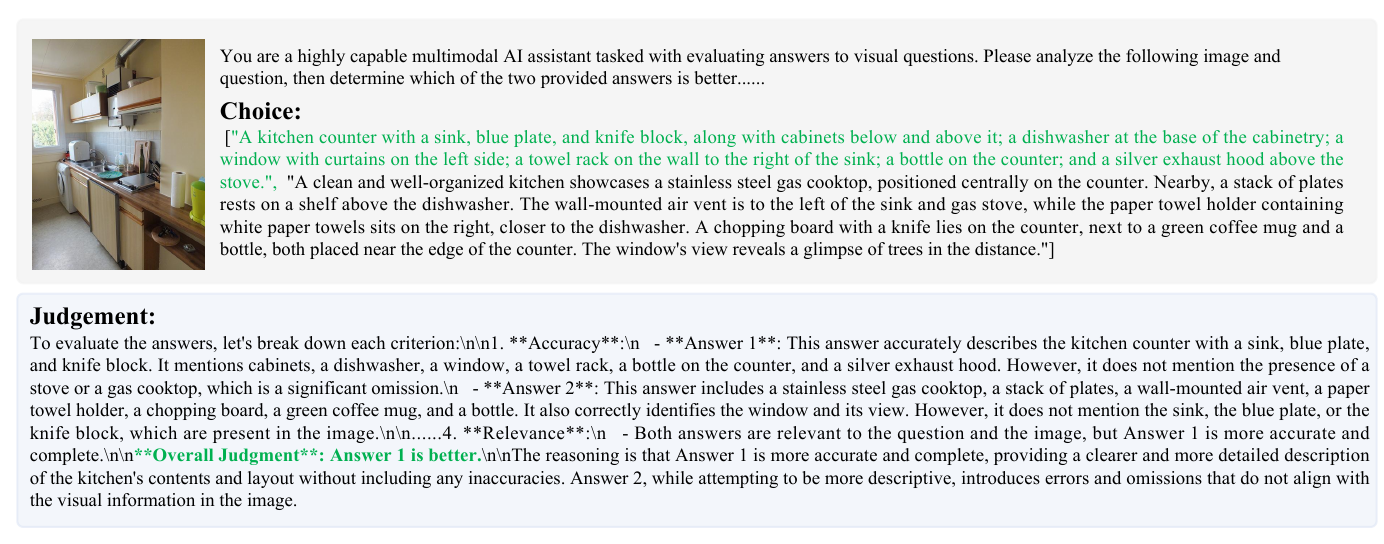}
    \end{center}
    \vspace{-18pt}
    \caption{\small \textbf{Reward Reasoning Case.} The example is taken from VL-RewardBench~\citep{li2025vlrewardbench}.
    }
    \vspace{-12pt}
    \label{fig:reward_case}
\end{figure}

\begin{figure}[t]
    \begin{center}
    \includegraphics[width=1.\linewidth]{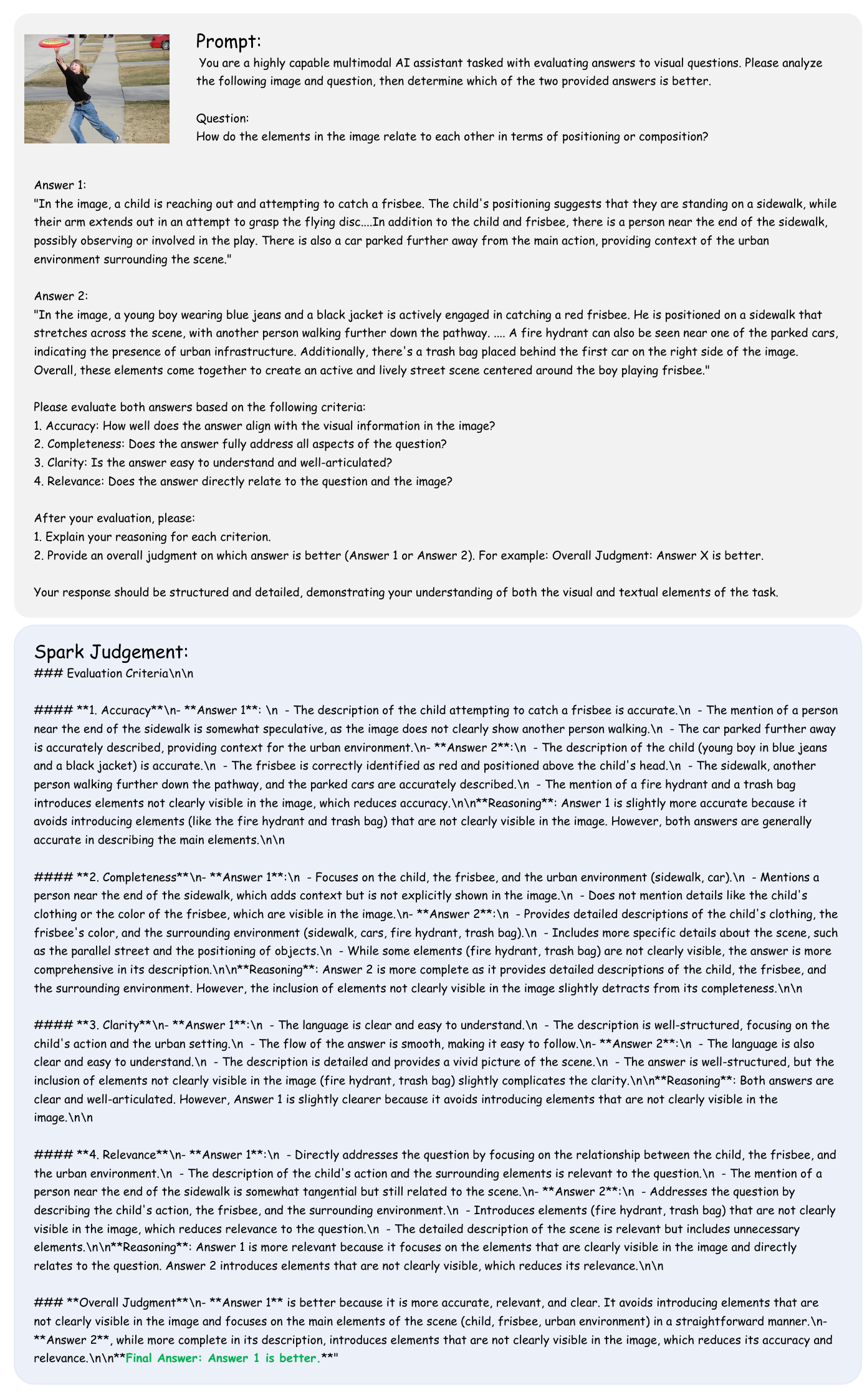}
    \end{center}
    \caption{\small \textbf{VL-RewardBench Inference Case 1.}}
    \vspace{-12pt}
    \label{fig:vl_case_sup1}
\end{figure}

\begin{figure}[t]
    \begin{center}
    \includegraphics[width=1.\linewidth]{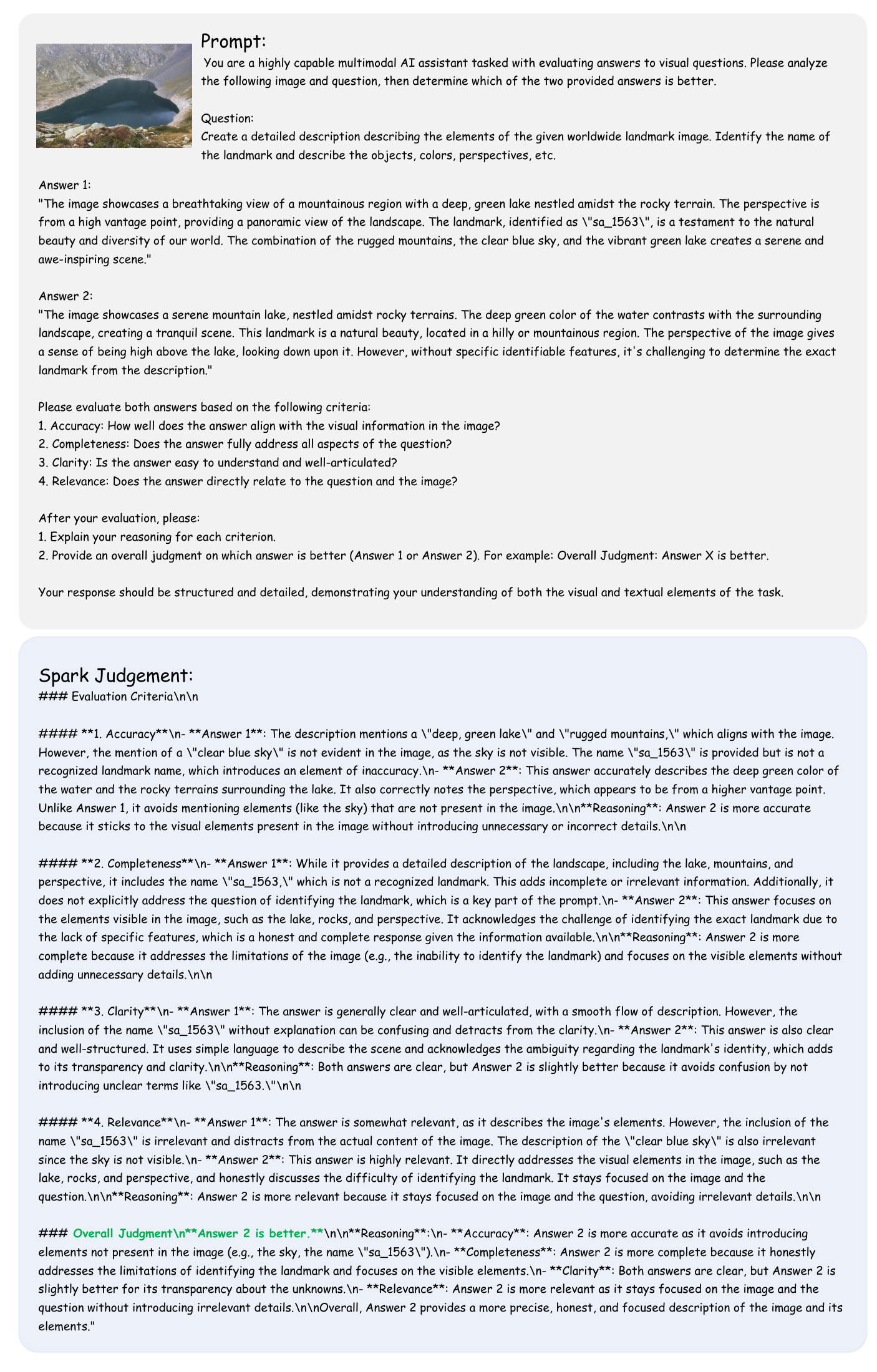}
    \end{center}
    \caption{\small \textbf{VL-RewardBench Inference Case 2.}}
    \vspace{-12pt}
    \label{fig:vl_case_sup2}
\end{figure}

Furthermore, in Fig.~\ref{fig:reward_case}, Fig.~\ref{fig:vl_case_sup1} and Fig.~\ref{fig:vl_case_sup2}, we showcase cases from VL-RewardBench. These examples demonstrate that the judgment ability acquired by our method in mathematical tasks can directly transfer and generalize to broader visual domains.

\subsection{The Evolving Landscape of Large Vision Language Models}\label{sec:appendix_landscape}
Multimodal Large Language models (MLLMs) have progressed rapidly, moving from early vision–language instruction tuning~\citep{liu2023visual,chen2024openllavanext}, fueled by the rapid growth of large-scale multimodal datasets and benchmarks~\citep{mmbench, liu2024mmdu, xing2025scalecap}, to advanced systems like Qwen2.5-VL~\citep{bai2025qwen25vl}, and InternVL~\citep{cai2024internlm2}. These models integrate vision encoders with LLM backbones, extending capabilities to document parsing, video understanding~\citep{wei2025videorope}, and multi-image understanding~\citep{liu2024mia}. Recent progress also highlights efficiency improvements through model compression and acceleration~\citep{xing2024pyramiddrop}, as well as the integration of retrieval-augmented generation (RAG) to enhance grounding and knowledge coverage in multimodal tasks~\citep{liu2024rar}.

Training and alignment now hinge on preference and reinforcement learning methods. Direct Preference Optimization (DPO)~\citep{dpo} has become the standard for stable, large-scale alignment~\citep{ouyang2022training,liu2024mia}. Reinforcement approaches such as RLHF and GRPO add verifiable reward signals for math~\citep{qwen25math}, logic, tool use~\citep{jin2025searchr1,liu2025visualrft} and agent~\citep{sun2025coda,sun2025seagent}. The prevailing recipe is hybrid: use DPO for broad preference alignment, or apply RL with verifiable feedback to sharpen reasoning.

\end{document}